\documentclass[11pt]{article}

\usepackage[final]{acl}
\usepackage{times}
\usepackage{latexsym}
\usepackage[T1]{fontenc}
\usepackage[utf8]{inputenc}
\usepackage{microtype}
\usepackage{inconsolata}

\usepackage{amsmath}
\usepackage{amssymb}
\usepackage{siunitx}

\usepackage{graphicx}
\usepackage{float}
\usepackage{placeins}

\usepackage{booktabs}
\usepackage{multirow}
\usepackage{url}

\ifdefined\linenumberfont\renewcommand{\linenumberfont}{\normalfont\tiny\color{lightgray}}\fi

\title{HydraLLM: Adaptive Width Slicing \\ for Elastic Text and Vision-Language Models}

\author{Oscar Kemme\textsuperscript{1}, Janek Haberer\textsuperscript{2}, Ali Hojjat\textsuperscript{1,2}, Olaf Landsiedel\textsuperscript{1,2,3} \\
  \textsuperscript{1}Hamburg University of Technology (TUHH), Germany \\
  \textsuperscript{2}Kiel University, Germany \\
  \textsuperscript{3}UNU-INWEH, Germany \\
  \texttt{\{oscar.kemme,ali.hojjat,olaf.landsiedel\}@tuhh.de} \\
  \texttt{janek.haberer@cs.uni-kiel.de}}

\begin{document}
\maketitle

\begin{abstract}
LLMs and LVLMs are increasingly deployed across heterogeneous hardware, ranging from data-center GPUs to resource-constrained edge devices. Their fixed-size architectures limit the ability to reuse a single model across deployment settings with different compute, latency, and model-size targets. We introduce HydraLLM\footnotemark, an elastic MQA-based architecture. Its number of active attention heads determines the effective widths of the embeddings, attention projections, feed-forward layers, and the output head, forming a nested family of subnetworks within a single parameter set. We introduce a static-to-elastic conversion pipeline that, starting from a pre-trained model checkpoint, includes importance-driven weight reordering and stochastic subnetwork training with knowledge distillation. We extend this principle to multimodal inference through an elastic projection module whose output width follows the active text width. HydraLLM covers a wider range of active parameter configurations than FFN-only elastic architectures. With 75\% of heads active, HydraLLM achieves 2.2 percentage points higher Trivia\-QA accuracy than a state-of-the-art nested baseline configuration which has 17.25\% more parameters. With 50\% of heads active, our multimodal model achieves a CLIPScore comparable to that of the 7$\times$ larger Gemma 3 4B model. With only 25\% of heads active, HydraLLM increases token-generation throughput by 75.3\% over the full-width configuration.
\end{abstract}
\footnotetext{The source code is available at \url{https://github.com/tuhh-ncps/hydrallm}.}

\section{Introduction}
\begin{figure}[t]
    \centering
    \includegraphics[width=0.66\linewidth]{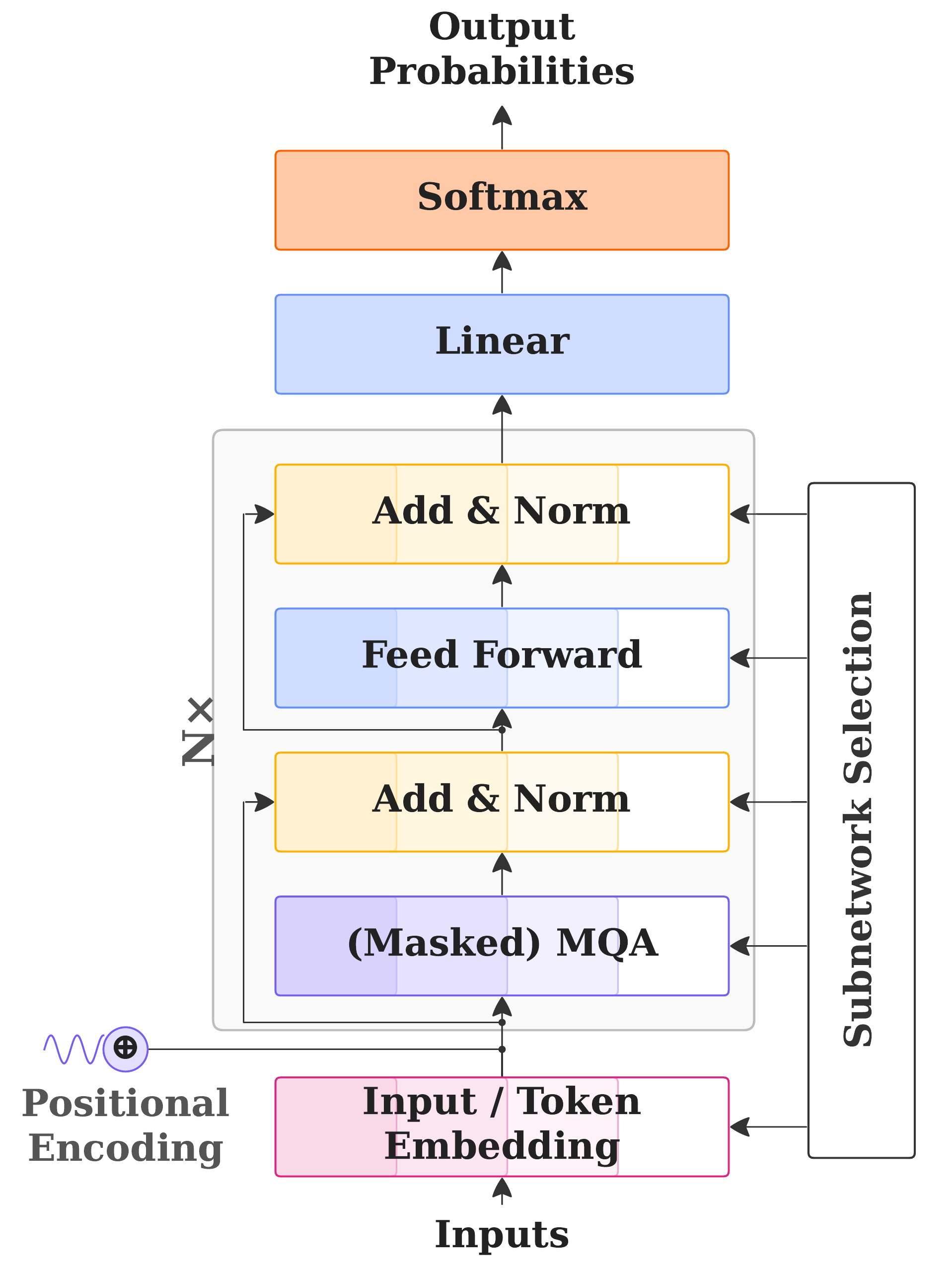}
    \caption{HydraLLM width slicing. The number of active heads determines the embedding dimension, the attention projection, and the feed-forward network width, producing nested subnetworks from a single parameter set.}
    \label{fig:hydravit_all_widths}
    \vspace{-10pt}
\end{figure}

\begin{figure*}[t]
    \centering
    \includegraphics[width=\textwidth]{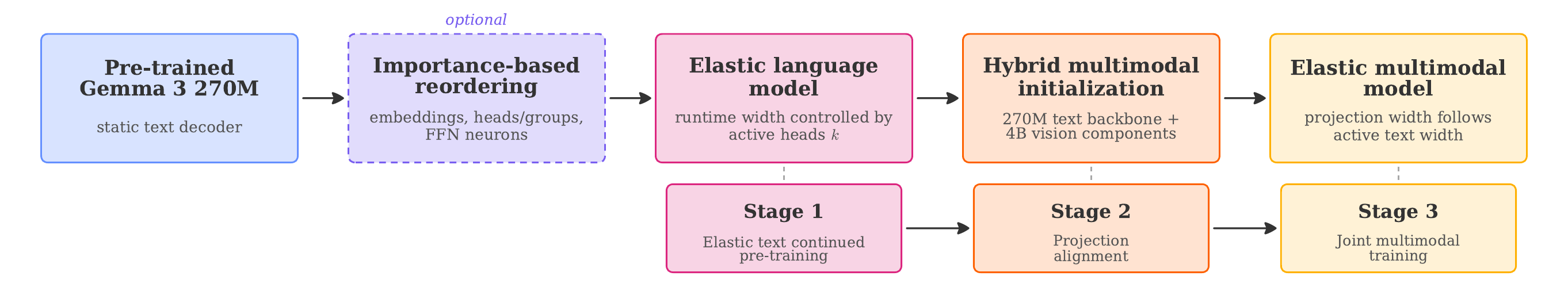}
    \caption{Pipeline overview. Starting from a static text checkpoint, such as Gemma 3 270M, the sliced axes can first be reordered to improve prefix quality. The resulting checkpoint is then trained as an elastic text model whose runtime width is controlled by the active number of query heads $k$. Finally, we combine the elastic text decoder with, for example, Gemma 3 4B vision components to form a multimodal model whose projection output follows the same active width.}
    \label{fig:pipeline_overview}
\end{figure*}

Large Language Models (LLMs) and Large Vision-Language Models (LVLMs) underpin applications such as document understanding~\cite{molmo_pixmo_deitke2024}, conversational assistants~\cite{openai_chatgpt_overview}, and agent systems~\citep{google_gemini_assistant}. 
Transformer-based architectures~\cite{attentionneed_vaswani2017} exhibit robust scaling, driving the creation of larger foundation models. At the same time, their deployment spans heterogeneous hardware from data-center GPUs to resource-constrained edge devices with different latency, compute, and model-size requirements.
Training separate models for each configuration is computationally prohibitive, as each variant must be trained independently from scratch.
Additionally, conventional transformer-based models are inherently static at inference time, which means their parameters are fixed after training, limiting their ability to target heterogeneous systems.
Recent elastic and nested methods address this by training multiple models as one, but most focus on a single component such as the feed-forward network (FFN) for size reduction or require a multi-step extraction process of fixed model checkpoints~\citep{hydravit_haberer2024, matformer_devvrit2024, attentionneed_vaswani2017, onceforall_cai2020}. 
In this paper, we introduce HydraLLM\footnotemark[\value{footnote}], a unified approach to adaptive slicing in LLMs and LVLMs by merging elastic architectural principles with efficient compression techniques while leveraging an already pre-trained static LLM.

We show that HydraViT-style~\citep{hydravit_haberer2024} end-to-end adaptive slicing transfers successfully to decoder-only LLMs, producing a single checkpoint that remains effective across several runtime width budgets. With 75\% of heads active, HydraLLM improves Trivia-QA~\citep {triviaqa_joshi2017} accuracy by 2.2 percentage points over a baseline configuration that has 17.25\% more parameters. With 50\% of heads active, the multimodal model reaches a CLIPScore of 66.9, surpassing the 65.3 score of the 7$\times$ larger Gemma 3 4B~\cite{gemma3_gemmateam2025} model. With only 25\% of heads active, importance-based reordering improves post-training perplexity by 16.2\%. This 25\% configuration increases token-generation throughput by 75.29\% compared to the full model.
In summary, we contribute:
\begin{list}{$\bullet$}{\leftmargin=1em\itemsep=0.5ex}
    \item \textbf{End-to-end slicing for decoder-only LLMs.} HydraLLM exposes a single head-count control that selects consistent prefixes of the embedding, attention, FFN, and output dimensions, producing nested decoder subnetworks from one checkpoint.
    \item \textbf{Unified training pipeline for elastic subnetworks.} We introduce a single training recipe to convert a static pre-trained LLM to an adaptively sliceable LLM while remaining competitive on text-only benchmarks.
\linebreak
    \item \textbf{Elastic multimodal projection for LVLMs.} We extend adaptive slicing to LVLMs by employing an elastic projection module that maps visual features into text embedding spaces of varying dimensionality.
\end{list}

\section{Related Work}

We discuss related work on nested representations and elastic architectures, transformer compression and submodel extraction, and projection-based multimodal models. We focus on how these approaches differ from HydraLLM's end-to-end runtime slicing of a pre-trained LLM and its extension to an elastic multimodal projection.

\citet{matryoshka_kusupati2024} introduce multi-granularity, coarse-to-fine representations for embedding models.
The same matryoshka nesting principle reappears in later work introducing elasticity within other transformer components.
\citet{onceforall_cai2020} propose the Once-For-All (OFA) methodology, decoupling architecture search from training by progressively shrinking the convolutional neural network (CNN) kernel size, channel width, and model depth.
OFA targets the search and extraction of specialized CNN architectures for different deployment constraints. In contrast, HydraLLM does not perform architecture search or progressive shrinking: it converts a pre-trained LLM into one elastic checkpoint whose embedding, attention, FFN, and output widths can be scaled jointly at runtime through the active head count.
\citet{matformer_devvrit2024} propose MatFormer, a natively elastic transformer that introduces a Matryoshka-like nested structure specifically \emph{only} inside the FFN block. 
\citet{hydravit_haberer2024} target embeddings, attention heads, and FFN layers for their nested ViTs.
Training employs stochastic subnetwork sampling by drawing one head count $k$ per step and updating only the corresponding subnetwork, which biases optimization toward earlier heads, implicitly ordering them by importance~\citep{hydravit_haberer2024}. 
We apply the same training and architectural approach to LLMs and extend it to the vision domain.
\citet{flextron_cai2024} and \citet{nemotron_elastic_taghibakhshi2025} extend the nested approach in the attention and FFN components with routers in their FLEXTRON and Nemotron Elastic models.
Both approaches are text-only. FLEXTRON slices only attention heads and FFN, and Nemotron Elastic targets an architecture distinct from our multi-query attention (MQA)~\citep{mqa_shazeer2019} transformer.
\citet{compactlanguagemodelspruning_muralidharan2024} design a static submodel extraction approach by applying iterative structured pruning on a pre-trained model, creating MINITRON.
They obtain separate models that are not suitable for elastic inference, like ours.

ViTs~\citep{imageworth16x16words_dosovitskiy2021} adapt the transformer encoder to images by splitting an image into non-overlapping patches, flattening each patch, and projecting it into the model's embedding space to obtain patch embeddings.
For multimodal models, \citet{llava_liu2024} connect a pre-trained CLIP~\citep{clip_radford2021} vision encoder to an open-source LLM via a projection layer that maps visual features into the LLM's text embedding space, producing so-called soft tokens. They merge the soft tokens with text tokens and process them jointly through self-attention.
\citet{molmo_pixmo_deitke2024} build on these concepts with the Molmo family, while \citet{smolvlm_marafioti2025} propose SmolVLM as a series of compact multimodal models. 
\citet{gemma3_gemmateam2025} release Gemma 3 with a tailored SigLIP~\citep{siglip_zhai2023} vision encoder.
We follow the training and architectural principles of these multimodal models and utilize insights from their training approaches.
However, we employ the elastic LLM and multimodal projection.

\section{HydraLLM: Architecture and Pipeline}
\label{sec:pipeline-architecture}
This section introduces HydraLLM's design and the pipeline for converting a static pre-trained LLM into an elastic text and multimodal model.
Specifically, we describe the text architecture and width-control in Section~\ref{sec:text-backbone-width-control}, the static-to-elastic conversion procedure including reordering and the optimization objective details in Section~\ref{sec:static-to-elastic-conversion}, and the multimodal assembly in Section~\ref{sec:multimodal-assembly}.
Our pipeline starts from a static pre-trained text checkpoint and converts it into an elastic decoder whose active width is controlled at runtime; see Figure~\ref{fig:pipeline_overview}. 
We then extend the same elasticity principle to the multimodal setting with an elastic projection module.

\subsection{Runtime Width-Control}
\label{sec:text-backbone-width-control}
Hydra\-LLM adapts HydraViT's~\citep{hydravit_haberer2024} head-stacking principle to decoder-only language models by using the number of active query heads \(k\) as a single runtime width-control parameter; see Figure~\ref{fig:hydravit_all_widths}.
A subnetwork keeps the first \(k\) query heads together with the corresponding prefixes of the embedding channels, output projection, and FFN neurons.
Each sliced dimension scales linearly with the number of heads \(k\) in relation to the total number of query heads \(H_{\mathrm{q}}\): the active embedding width is \(d_{E,\mathrm{active}} = k/H_{\mathrm{q}} \times d_E\), with the FFN dimensions following the same proportion.
This produces a strict nesting of subnetworks that share parameters across configurations.
At inference, selecting \(k\) computes only the corresponding prefix, so one checkpoint serves all width budgets.
We target MQA~\citep{mqa_shazeer2019} in our design, in which all query heads share a single key-value head.

\subsection{Static-to-Elastic Conversion}
\label{sec:static-to-elastic-conversion}

A static pre-trained checkpoint is not immediately suitable for prefix slicing. 
In an unmodified checkpoint, units of a component are usually not sorted by their importance.
Since prefix slicing always retains the first units and discards later units, na\"{i}ve slicing can remove important parameters and severely degrade smaller subnetworks.
HydraLLM employs a two-stage procedure to convert static models into elastic ones. First, the pipeline reorders the checkpoint offline, moving important units toward the beginning of each sliced dimension so that prefix slicing retains the most informative parameters.
Second, it optimizes the checkpoint using stochastic subnetwork training: during each training step, the pipeline samples a single $k$ and updates only the corresponding active subnetwork.
Since earlier units are active for more sampled values of $k$, they receive proportionally more gradient updates, reinforcing the importance ordering and adapting the shared parameters to all configured widths~\citep{nemotron_elastic_taghibakhshi2025,compactlanguagemodelspruning_muralidharan2024,flextron_cai2024,hydravit_haberer2024,matformer_devvrit2024}.

\subsubsection{Importance Scoring and Reordering}
\label{sec:importance-reordering}

The reordering step occurs before any training and requires the definition of importance.
We compare three importance measures: 
Activation magnitude~\cite{compactlanguagemodelspruning_muralidharan2024}, variance-weighted consumer strength (VxC) adapted from \citet{variance_weinstein2022}, 
and CETT~\citep{cett_zhang2024}, which scores units by their normalized contribution to the component output and therefore aligns well with prefix slicing.
We estimate unit importance from activation statistics collected during a forward pass over a calibration text corpus.
After computing the importance scores, we permute the corresponding tensors so that important units appear earlier along the sliceable dimensions: 
For embedding-width reordering, we apply one global permutation to the token embedding channels and propagate it consistently through the RMS\-Norm weights, the input columns of the Q/K/V projections, the output rows of the attention output projection, the input columns of the FFN gate and up projections, the output rows of the FFN down projection, the final normalization layer, and, if untied, the language-model head.
For attention reordering, we permute the query heads and the corresponding attention-output input blocks.
For FFN reordering, we permute FFN neurons by moving the rows of the gate and up projections and the corresponding columns of the down projection.
This reordering step preserves the function of the full model.

\subsubsection{Optimization Objective}
\label{sec:optimization-objective}

After reordering, we train the model.
The training objective we employ is a combination of autoregressive language modeling, logit regularization, and teacher supervision.
The primary task loss, \(\mathcal{L}_{\mathrm{CE}}\), is standard next-token cross-entropy.
We additionally add a \(z\)-loss term \(\mathcal{L}_z\), weighted by \(\lambda_z\) \citep{palm_chowdhery2022}, and logit-based knowledge distillation \citep{distilling_knowledge_hinton2015}.
Let \(z_t\) and \(z_s\) denote the teacher and student logits, respectively, and let \(p_t^\tau = \mathrm{softmax}(z_t/\tau)\) and \(p_s^\tau = \mathrm{softmax}(z_s/\tau)\) denote their temperature-scaled output distributions. The training objective is:
\begin{equation}
    \label{eq:training-objective}
    \begin{aligned}
        \mathcal{L}_{\mathrm{KD}}
        &= \tau^2 \mathrm{KL}\!\left(p_t^\tau \,\|\, p_s^\tau\right), \\
        \mathcal{L}
        &= \mathcal{L}_{\mathrm{CE}}
        + \lambda_z \mathcal{L}_z
        + \lambda_{\mathrm{KD}} \mathcal{L}_{\mathrm{KD}}
    \end{aligned}
\end{equation}
We keep the teacher model frozen throughout training. 

\subsection{Elastic Multimodal Assembly}
\label{sec:multimodal-assembly}

\begin{figure}[t]
    \centering
    \includegraphics[width=1\linewidth]{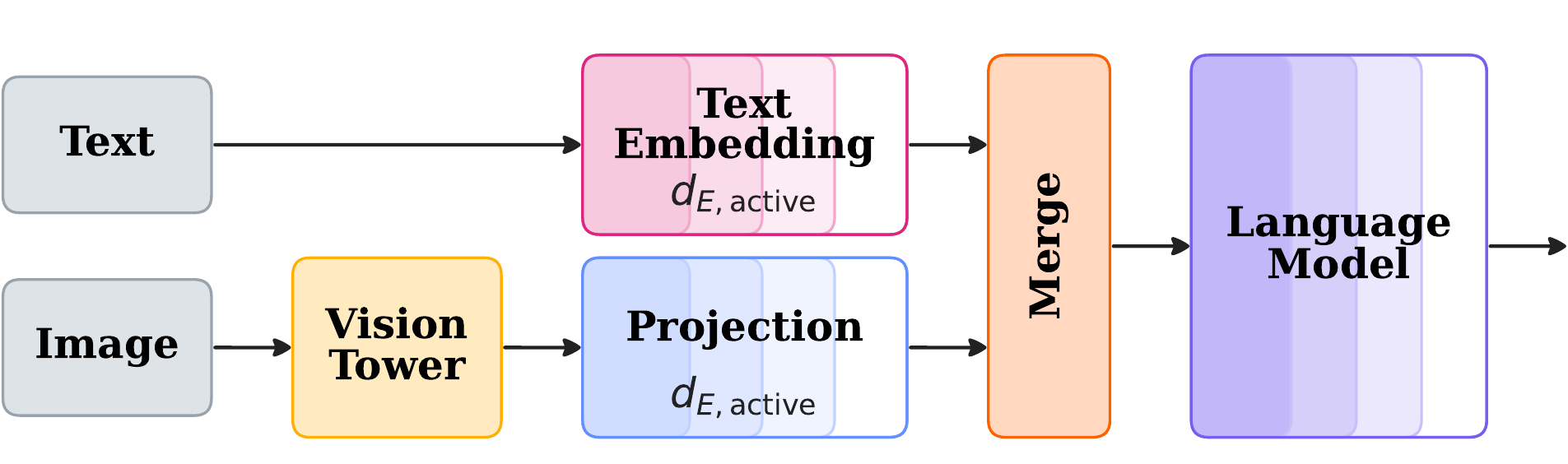}
    \caption{Elastic LVLM architecture. The projection output width follows the active text width.}
    \label{fig:elastic_lvlm_architecture}
\end{figure}

We construct the multimodal variant of HydraLLM by combining the language decoder with a static pre-trained SigLIP-based vision encoder~\citep{gemma3_gemmateam2025,siglip_zhai2023}.
The key difference from a conventional projection-based LVLM is that HydraLLM also makes the projection elastic. Let \(d_P\) denote the projection output width. 
We enforce \(d_{P,\mathrm{active}} = d_{E,\mathrm{active}}\), so the same runtime parameter \(k\) controls both the active text decoder width and the projection output width; see Figure~\ref{fig:elastic_lvlm_architecture}. 
This creates a single elasticity axis across modalities: 
selecting a smaller text subnetwork also selects a correspondingly smaller visual projection output.

\section{Experiment Setup}

We describe our training data and evaluation protocol in the following subsections~\ref{sec:training-recipe} and~\ref {sec:evaluation_protocol}, respectively.

\subsection{Training Recipe}
\label{sec:training-recipe}

We base our experiments on Google Gemma 3 270M~\cite{gemma3_gemmateam2025}, with \(k \in \{1,2,3,4\}\); see Appendix~\ref{sec:runtime_config}.
For text-only experiments, we use the frozen full-width Gemma 3 270M as the teacher model.

Most text experiments train on a subset of FineWeb-Edu 350BT \citep{lozhkov2024fineweb-edu}. 
We concatenate text samples with an \texttt{<eos>} separator and split the text into fixed-length token blocks.
We employ non-uniform runtime width sampling of $k$, placing additional ($\times$3) emphasis on the full-width model compared to the submodels; see Appendix~\ref{sec:appendix-hyperparameters}.
In contrast to the packed text pre-training setup, the optional text fine-tuning stage uses text samples as they appear, without fully packing the sequence length; see Appendix~\ref{sec:text-recovery-mm}.

For multimodal training, we use the elastic HydraLLM Gemma 3 270m model combined with the vision encoder and projection module initialized from the Gemma 3 4B multimodal checkpoint.
The teacher for this training is the static and frozen Gemma 3 4B model.
We employ uniform head sampling and separate learning rates for the text backbone, the projection module, and, optionally, the vision encoder.
We configure separate peak learning rates and number of warmup steps for the text backbone, projection and vision encoder; see Appendix~\ref{sec:appendix-hyperparameters}.
Multimodal training samples from PixMoCap, the COCO Recaptioned dataset, LLaVA Pretrain 558K~\citep{liuhaotian_llava_pretrain}, PixMo Ask Model Anything, PixMo Cap-QA, CoSyn-400K, and text-only data from Wikipedia \citep{wikimedia_wikipedia} and FineWeb-Edu.
See Appendix~\ref{sec:dataset-compositions} for detailed composition. Projection alignment utilizes 128{,}000 captioning samples from the aforementioned captioning datasets.

\subsection{Metrics and Evaluation Datasets}
\label{sec:evaluation_protocol}

\begin{table}[tb]
\centering
\small
\setlength{\tabcolsep}{5pt}
\begin{tabular}{lcccc}
\toprule
\multirow{2}{*}{Recipe} & \multicolumn{4}{c}{Perplexity $\downarrow$} \\
\cmidrule(lr){2-5}
 & H1 & H2 & H3 & H4 \\
\midrule
no reordering & 168.6M & 1.1M & 41.3k & 49.0 \\
mag-cett-cett & \underline{37.2k} & \underline{2.43k} & \underline{1.22k} & 49.0 \\
mag-mag-mag & 8.45M & 378.9k & 1.54k & 49.0 \\
vxc-vxc-vxc & 38.0M & 4.87M & 60.4M & 49.0 \\
\bottomrule
\end{tabular}
\caption{Perplexity of \textbf{reordered} checkpoints measured on samples of plays by William Shakespeare. Magnitude for embeddings combined with CETT for heads and FFN yields the strongest subnetwork perplexities, while VxC consistently hurts.}
\label{tab:ppl_results}
\end{table}

We employ a mix of metrics and benchmarks to evaluate models and derive conclusions.
For text evaluation, we report ARC-Easy~\citep{arc_clark2018} and TriviaQA~\citep{triviaqa_joshi2017}.
We evaluate ARC-Easy with both raw accuracy and answer-length-normalized accuracy.
We evaluate TriviaQA with five-shot prompting and exact-match scoring. 
We measure perplexity~\citep{jm3} on samples of plays by William Shakespeare~\citep{balbay2024calibration} as a general-purpose indicator of language modeling quality.

For multimodal evaluation, we report CLIPScore \citep{CLIPScore,clip_radford2021} on COCO validation images and multiply the score by 100 for readability.
We additionally use qualitative captioning examples to inspect generation quality across different head configurations.
The standard error on text- and CLIPScore-benchmark results denotes standard errors over evaluation examples.

We measure throughput with \texttt{llama-bench} from \texttt{llama.cpp} \citep{llama_cpp}, with prompt-processing of 2048 tokens and autoregressive generation of 512 tokens on an NVIDIA RTX PRO 4000.
This setup allows measuring prompt-processing and token generation speed improvements.

\section{Evaluation}
\label{sec:evaluation-protocol}

We evaluate the static-to-elastic conversion pipeline in terms of reordering in Section~\ref{sec:reordering}, compare HydraLLM with MatFormer-style~\citep{matformer_devvrit2024} FFN slicing in terms of accuracy and throughput in Section~\ref{sec:hydraLLM_matformer_comparison}, and assess the quality of multimodal extensions in Section~\ref{sec:multimodal_pipeline_results}.

\subsection{Effect of Importance Reordering}
\label{sec:reordering}

To determine the effect of reordering, we measure perplexity for each head configuration and compare it to the unmodified pre-trained model in Table~\ref{tab:ppl_results}.
We denote the $k$-head configurations as H1-H4.
Reordering clearly improves the perplexity of submodels when utilizing the magnitude metric for embeddings combined with CETT for heads and FFN, or the magnitude metric for all components. Employing only VxC provides an improvement solely for the H1 configuration.

After reordering, we train the reordered models for 20{,}000 steps ($\sim$1B tokens) and measure their perplexities again.
\begin{table}[tb]
\centering
\small
\setlength{\tabcolsep}{5pt}
\begin{tabular}{lcccc}
\toprule
\multirow{2}{*}{Recipe} & \multicolumn{4}{c}{Perplexity $\downarrow$} \\
\cmidrule(lr){2-5}
 & H1 & H2 & H3 & H4 \\
\midrule
no reordering & 123.9 & 77.4 & 56.7 & 45.2 \\
mag-cett-cett & \underline{106.6} & \underline{71.6} & \underline{54.4} & 44.3 \\
mag-mag-mag & 110.1 & 71.9 & 54.4 & 44.8 \\
vxc-vxc-vxc & 118.1 & 71.7 & 55.0 & \underline{44.1} \\
\bottomrule
\end{tabular}
\caption{Perplexity of \textbf{reordered and trained} (20k steps) checkpoints measured on samples of plays by William Shakespeare. Magnitude for embeddings combined with CETT for heads and FFN yields the strongest subnetwork perplexities.}
\label{tab:ppl_results_after_reordering}
\end{table}
After training, perplexity improves for all heads, regardless of the starting point; see Table~\ref{tab:ppl_results_after_reordering}.
However, the CETT importance metric for heads and FFN with magnitude importance for the embeddings provides the best results for the submodels. 
VxC slightly edges out CETT on the four-head configuration.
For H4 of the not reordered model, continued training reduces perplexity from 49.0 to 45.2. However, this improvement does not consistently transfer to downstream benchmarks.
Figure~\ref{fig:arc_vs_size} and Figure~\ref{fig:trivia_vs_size} from the next Section~\ref{sec:hydraLLM_matformer_comparison} also show the ARC-Easy and Trivia\-QA accuracy of the pre-trained and the no-reordering model trained for 20{,}000 steps.
The benchmarks actually show a lower accuracy compared to the static baseline. We provide more benchmark results in Appendix~\ref{sec:appendix-reordering-ablation}.

\subsection{HydraLLM vs. MatFormer}
\label{sec:hydraLLM_matformer_comparison}

\begin{figure}[t]
    \centering
    \includegraphics[width=\linewidth]{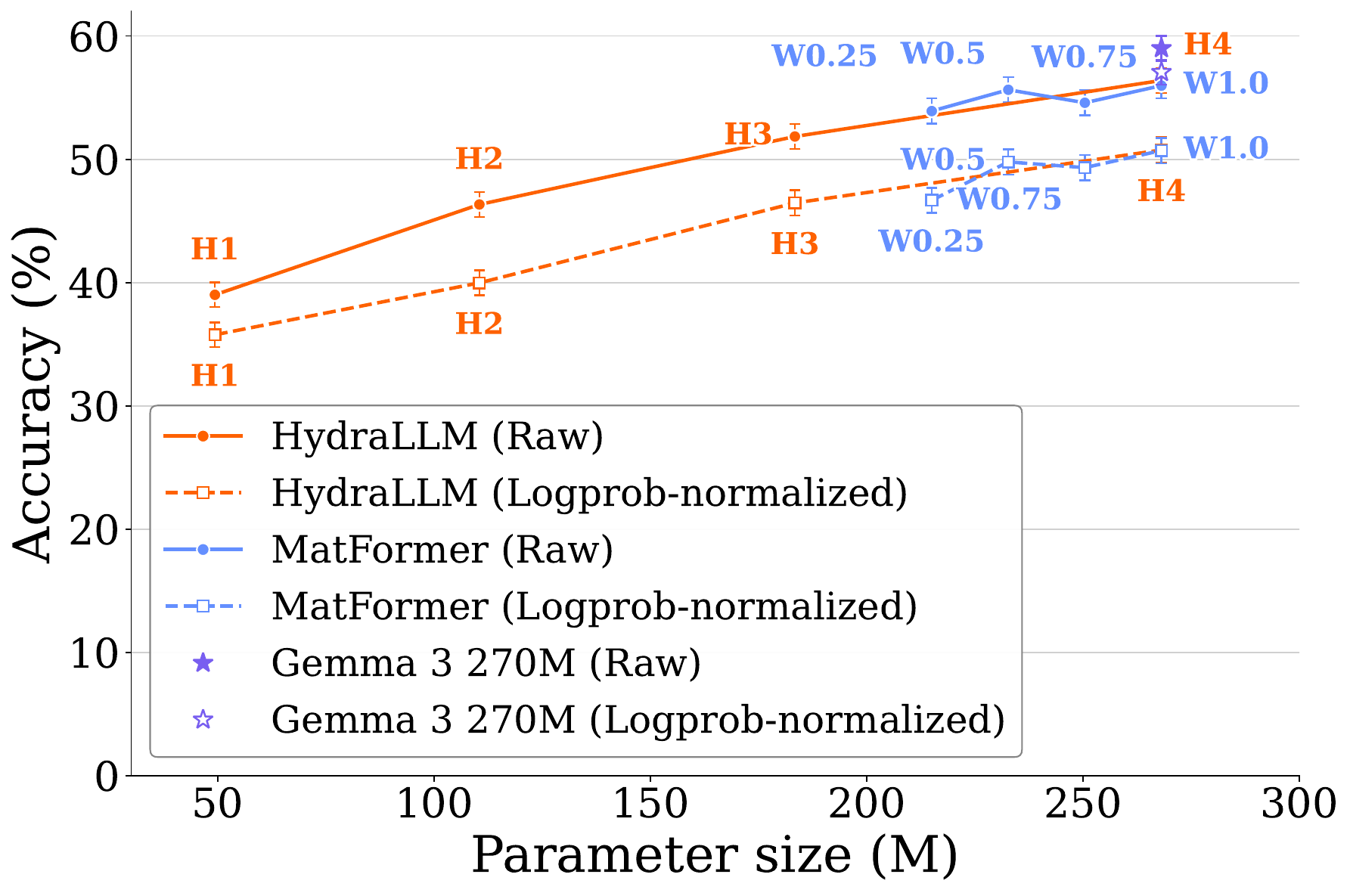}
    \caption{ARC-Easy accuracy as a function of active parameter size for HydraLLM and MatFormer configurations. HydraLLM covers a wider active parameter range than MatFormer while offering competitive accuracies.}
    \label{fig:arc_vs_size}
\end{figure}

\begin{figure}[t]
    \centering
    \includegraphics[width=\linewidth]{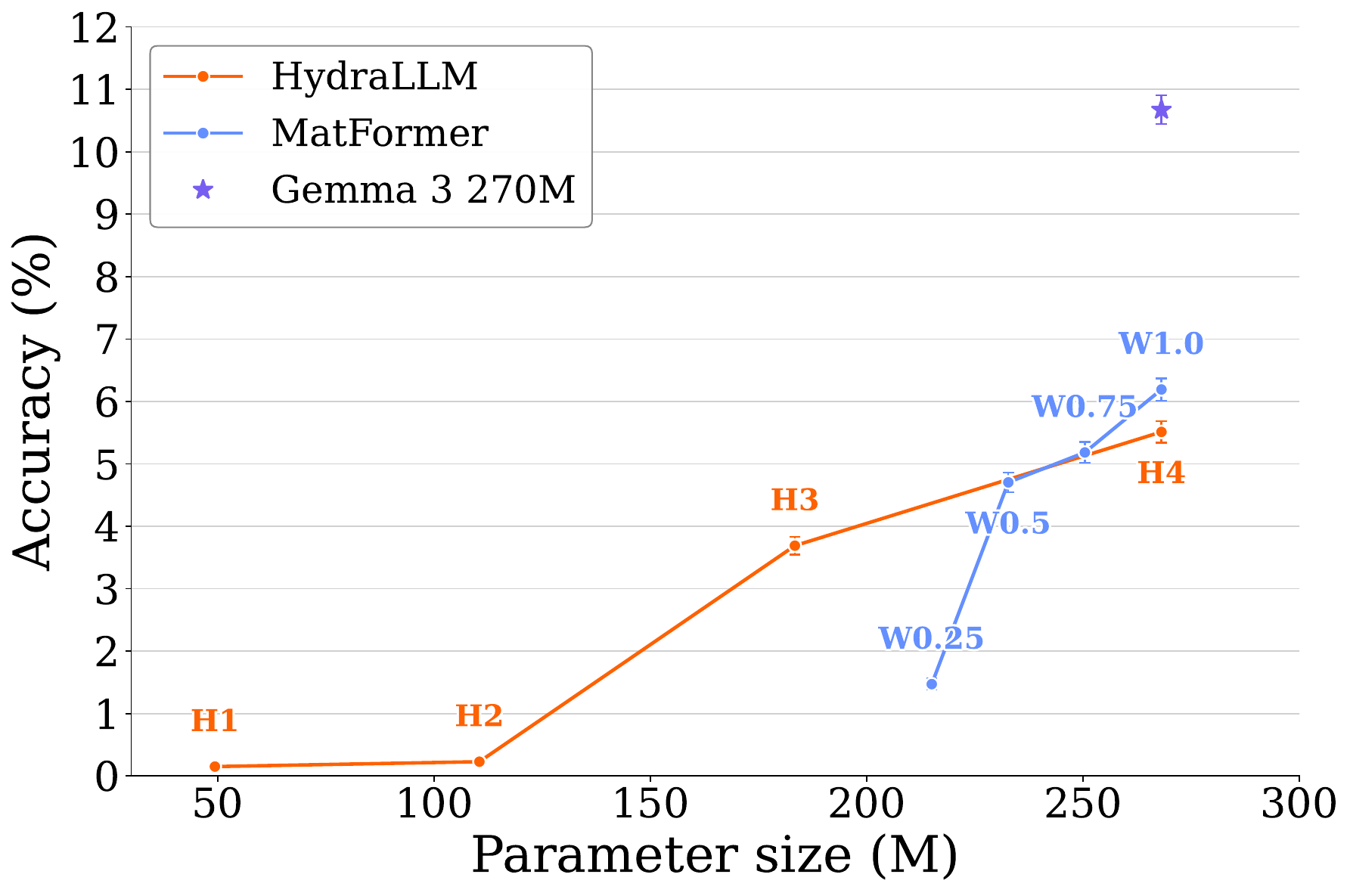}
    \caption{Trivia\-QA accuracy as a function of active parameter size for HydraLLM and MatFormer configurations. HydraLLM is stronger at H3 clearly outperforming the similarly sized MatFormer W0.25, while MatFormer surpasses HydraLLM at the full configuration.}
    \label{fig:trivia_vs_size}
\end{figure}

We compare HydraLLM to MatFormer-style FFN slicing, which we reimplement using the same Gemma 3 270M checkpoint, in terms of accuracy and runtime throughput.
The hyperparameters for both models are the same, apart from the width selection mechanism.
In Figures~\ref{fig:arc_vs_size} and~\ref{fig:trivia_vs_size}, we label the MatFormer FFN configurations as W0.25-W1.0, corresponding to the fraction of FFN neurons active. 
Because MatFormer only alters the FFN component, the smallest MatFormer configuration, W0.25, still exceeds the second-largest HydraLLM configuration, H3, by 17.25\% in parameter count. 
The raw ARC-Easy accuracy of W0.25 is slightly higher than that of H3, though not statistically significant. The log-probability-normalized accuracies are virtually the same.
The HydraLLM approach has a clear advantage with similar accuracies, but a wider parameter size coverage.
The ten-shot prompting HellaSwag~\citep{hellaswag_zellers2019} results lead to the same conclusion, therefore we provide them in Appendix~\ref{sec:ablations} only.

The results on Trivia\-QA differ. 
While the W0.5 and W0.75 MatFormer configurations interpolate smoothly between the H4 and H3 HydraLLM configurations in Figure~\ref{fig:trivia_vs_size}, H3 achieves 2.21 percentage points ($\times$2.51) higher accuracy than W0.25, although they are similar in size.
Conversely, W1.0 outperforms H4 by 0.68 percentage points.
The HydraLLM approach has a broader parameter spread, starting from 49.34M parameters, compared to MatFormer, starting from 215.01M parameters, while achieving competitive results with similar submodel sizes. 
However, submodels H1 and H2 only achieve close to 0\% accuracy on this benchmark.

\begin{table}[tb]
\centering
\small
\setlength{\tabcolsep}{6pt}
\begin{tabular}{lrrrr}
\toprule
\multirow{2}{*}{Config} & \multicolumn{2}{c}{HydraLLM} & \multicolumn{2}{c}{MatFormer} \\
\cmidrule(lr){2-3} \cmidrule(lr){4-5}
 & Prefill & Decode & Prefill & Decode \\
\midrule
H1 / W0.25 & 55,891.3 & 1,057.0 & 51,549.1 & 669.6 \\
H2 / W0.50 & 52,851.7 & 785.9 & 51,353.8 & 648.5 \\
H3 / W0.75 & 51,165.4 & 594.2 & 49,945.2 & 623.4 \\
H4 / W1.00 & 50,487.9 & 603.0 & 50,509.8 & 602.3 \\
\bottomrule
\end{tabular}
\caption{Throughput in tokens per second for HydraLLM and MatFormer variants. Prompt-processing of 2048 tokens. Decode 512 tokens. Prompt-processing throughput gains are modest for both methods, while token-generation scaling is much larger for HydraLLM than for MatFormer.}
\label{tab:throughput_table}
\end{table}

The prompt-processing improvements from smaller model configurations displayed in Table~\ref{tab:throughput_table} are very modest. 
The improvement from the H4 to H1 configuration ranges from 9.78\% to 11.61\%, and the improvement from W1.0 to W0.25 ranges from 1.03\% to 3.1\%. 
The Gemma 3 270M model comprises $\sim$170 million embedding parameters and $\sim$100 million transformer block parameters. 
Since the embeddings are mainly a lookup operation, they do not contribute much to throughput, but they account for a large portion of the parameter savings in the HydraLLM approach~\citep{gemma3_gemmateam2025, attentionneed_vaswani2017}.

For token generation, MatFormer shows consistent speed-ups with shrinking parameter count.
HydraLLM produces fewer tokens per second with the H3 configuration compared to H4, which we hypothesize stems from kernel-shape optimizations in llama.cpp.
The improvement in token generation speed from the largest to the smallest configuration is 75.29\% for HydraLLM and 11.17\% for MatFormer. 
The main takeaway is that prompt-processing gains are minor, while token-generation scaling is much greater, but does not scale one-to-one with parameter count.
The reported throughput gains therefore stem mostly from smaller embeddings and query, output, and FFN projections, because MQA~\cite{mqa_shazeer2019} prohibits key-value cache savings from width reduction.

\subsection{Multimodal Pipeline Results}
\label{sec:multimodal_pipeline_results}

\begin{figure*}[tb]
    \centering
    \includegraphics[width=\textwidth]{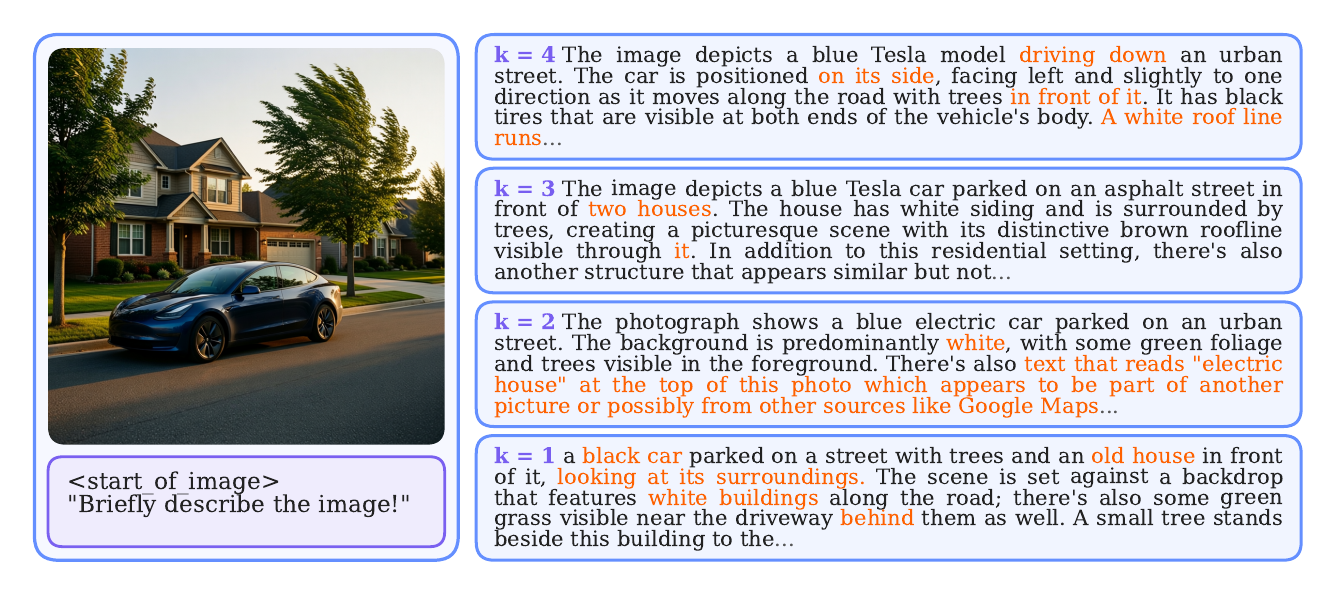}
    \caption{Qualitative captioning example. The elastic LVLM is prompted to describe an image of a car across the four head configurations ($k = 1$ to $k = 4$). Hallucinated or incorrect statements are highlighted in orange, illustrating the qualitative gap that persists alongside the encouraging CLIPScore trends.}
    \label{fig:example_figure_tesla}
\end{figure*}

The first stage in the multimodal training pipeline is aligning the projection with the text embedding space.
The projection-only block of Table~\ref{tab:mm_clip_recipes} shows that the untrained projection already produces a non-trivial CLIPScore~\citep{CLIPScore, clip_radford2021}  across all head configurations, indicating that it inherits useful structure from the Gemma 3 4B vision tower.
Training the projection in isolation yields only modest improvements over the untrained baseline.
During training, we observe a plateau in the loss after 300 to 350 steps, depending on the head configuration; see more details in Appendix~\ref{sec:appendix-align-projection-loss-curves}.
The rapid plateau indicates a lack of capacity in the projection layer, caused by the much smaller projection due to the lower text embedding dimension of the Gemma 3 270M model compared to Gemma 3 4B.
The projection has 25\% of the parameters of the Gemma 3 4B model and therefore possesses much less representational power~\citep{gemma3_gemmateam2025}.

Joint training of the projection with the text model substantially improves the CLIPScore; see the projection-plus-text-model block of Table~\ref{tab:mm_clip_recipes}.
The models trained from an already projection-pre-aligned checkpoint perform best across all head configurations, especially at H2, implying that aligning the projection before joint training of the projection and text model is beneficial.
When training all model parameters together, the all-parameters block shows that results are similar across different starting points, indicating that joint optimization can compensate for an inferior starting projection.

\begin{table}[tb]
\centering
\caption{CLIPScore ($\times100$) across head configurations for each multimodal training recipe. Bold marks the best score per head configuration.
\emph{non-elastic}: starting from pre-trained Gemma 3 270m and only head configuration $k{=}4$ is trained; \emph{elastic}: starting from elastic text model and all head configurations are trained stochastically; \emph{fine-tune}: an additional text-only fine-tuning step is applied before multimodal training; \emph{pre-aligned}: the projection is pre-aligned on captioning data before joint training; \emph{Gemma 3 4B}: the original frozen Gemma 3 4B multimodal model provided as a baseline.}
\label{tab:mm_clip_recipes}
\small
\begin{tabular}{lcccc}
\toprule
\textbf{Training Recipe} & \textbf{H1} & \textbf{H2} & \textbf{H3} & \textbf{H4} \\
\midrule
\multicolumn{5}{l}{\emph{No training}} \\
elastic             & 32.4 & 31.5 & 32.0 & 32.7 \\
Gemma 3 4B &  --  &  --  &  --  & 65.3 \\
\midrule
\multicolumn{5}{l}{\emph{Projection only (500 steps)}} \\
non-elastic & --  & --  & --  & 28.9 \\
elastic            & \textbf{38.2} & 31.4 & \textbf{31.1} & \textbf{33.5} \\
elastic, fine-tune & 34.8 & \textbf{34.2} & 31.0 & 32.1 \\
\midrule
\multicolumn{5}{l}{\emph{Projection + text model (2{,}500 steps)}} \\
elastic          & 46.3 & 53.9 & 58.5 & 58.8 \\
pre-aligned, non-elastic & --  & --  & --  & 62.5 \\
pre-aligned            & \textbf{47.4} & 56.9 & \textbf{61.3} & 61.0 \\
pre-aligned, fine-tune & 47.1 & \textbf{57.4} & 59.1 & \textbf{61.1} \\
\midrule
\multicolumn{5}{l}{\emph{All parameters (2{,}500 steps)}} \\
elastic           & 50.2 & 59.4 & \textbf{63.1} & \textbf{63.9} \\
pre-aligned            & 51.1 & \textbf{60.7} & 61.3 & 62.5 \\
pre-aligned, fine-tune  & \textbf{51.2} & 60.4 & 61.5 & 63.6 \\
\midrule
\multicolumn{5}{l}{\emph{From text fine-tune + pre-aligned (16{,}000 steps)}} \\
Projection + text model       & 59.1 & 65.4 & 65.1 & 68.0 \\
All parameters          & \textbf{62.6} & \textbf{66.9} & \textbf{68.6} & \textbf{69.6} \\
\bottomrule
\end{tabular}
\end{table}

Scaling the joint projection and text model training, as well as the full-parameter training, from 2{,}500 to 16{,}000 steps further closes the gap to the Gemma 3 4B reference; see the bottom block of Table~\ref{tab:mm_clip_recipes}.
The full-parameter trainable run matches or exceeds Gemma 3 4B at H2, H3, and H4 and reaches 62.6 at H1.
The results suggest that, on CLIPScore, the elastic model matches the vision-language capabilities of Gemma 3 4B across most subnetworks.
The cause for this lies in the response style of the Gemma 3 4B model: it often responds to captioning prompts with question-answer-style continuations rather than direct descriptive sentences, which reduces its CLIP similarity score.
CLIPScore is therefore sensitive to response style and does not, by itself, guarantee stable caption quality~\citep{CLIPScore}.

Given the strong CLIPScore results of HydraLLM compared to Gemma 3 4B, we evaluate our model with MMMU~\citep{mmmu_yue2024}.
Our models consistently achieve random accuracy of around $25\%$, whereas the Gemma 3 4B model reaches about $39\%$.
We illustrate the qualitative behavior with the captioning example in Figure~\ref{fig:example_figure_tesla}.
We prompt the model to describe an image of a car. Across all head configurations, the captions include hallucinated or incorrect statements about the scene.
Subject resolution, spatial relations, and object identity are unreliable: the four-head configuration hallucinates motion and orientation, the three-head and two-head configurations misidentify surrounding objects, and the single-head configuration misidentifies the car's color and introduces incorrect scene elements.
Hallucinations, weak grounding, and unstable object descriptions therefore persist across all head configurations of HydraLLM.
The elastic projection establishes feasibility, but the vision-to-language interface remains a suspected bottleneck and does not yet yield robust caption quality.

\section{Conclusion}
We introduce HydraLLM, an MQA-based~\cite{mqa_shazeer2019} elastic architecture that employs stochastic head stacking and end-to-end width slicing.
A single runtime parameter $k$ controls the active width of the decoder end-to-end, from token embeddings through attention and FFN projections to the output head, producing nested subnetworks from a single parameter set.
We extended the elasticity axis to LVLMs through an elastic projection module whose output width follows the active text width.

On text, magnitude-based reordering of embeddings combined with CETT-based~\citep{cett_zhang2024} reordering of heads and FFN units yields the strongest subnetworks, reducing perplexity by up to a factor of approximately 4,500 compared to no reordering.
However, the effect is reduced to at most 14\% after training. 
Knowledge distillation~\citep{distilling_knowledge_hinton2015} is crucial during training and improves TriviaQA exact-match accuracy at 75\% of active heads by 97.3\%.
HydraLLM covers a substantially wider parameter range at competitive accuracy compared to MatFormer-style~\citep{matformer_devvrit2024} FFN slicing. It achieves larger token-generation speed-ups of up to 75.3\%, but only modest prompt-processing improvements.

On the multimodal side, scaled training matches or exceeds the Gemma 3 4B CLIPScore~\citep{CLIPScore, clip_radford2021} of 65.3 by 2.5\%, 5.1\% and 6.6\% with two, three and four active heads respectively, confirming that the elastic text backbone transfers to the vision-language setting. However, the model remains near random accuracy on MMMU, and qualitative analysis reveals hallucinations and weak object grounding. These limitations point to clear avenues for future research: 
(i) replacing the constrained small single-layer projector with a larger or a multi-layer FFN projection, which would better align vision features with the elastic text backbone; 
(ii) longer and more diverse multimodal training schedules to improve grounding beyond captioning tasks; 
and (iii) combining HydraLLM's end-to-end slicing with MatFormer-style per-layer granularity for finer accuracy-compute control. We leave these directions for future work.

\section*{Acknowledgments}
We thank our anonymous reviewers for their insights and suggestions for improvement. This work has received funding from the Federal Ministry of Research, Technology and Space under the REAKT ISA project (grant no.~03DPC1415B) and from the Federal Ministry for Economic Affairs and Energy under the CAPTN X-FERRY project (grant no.~03SX612A).

\bibliography{references,custom}

\clearpage
\appendix

\section{Reproducibility Details}
\label{sec:implementation-details}

This section provides implementation and hardware details, runtime configurations, multimodal dataset composition, CLIPScore evaluation details, and training hyperparameters.

\subsection{Implementation and Hardware}

We build our implementation around PyTorch~\citep{pytorch_paszke2019} and Hugging Face Transformers~\citep{huggingfacestransformers_wolf2020}.
Accelerate~\citep{accelerate} and DeepSpeed~\citep{DeepSpeed_Rasley2020} support distributed training, and LightEval~\citep{lighteval} integrates the text benchmarks. CLIPScore follows \citet{CLIPScore}. 
When available, we use optional acceleration libraries such as FlashAttention~2 \citep{flashattention2_dao2023}, while ensuring a transparent fallback to the standard PyTorch and Transformers backends. 
Table~\ref{tab:packages} lists exact package versions.

\begin{table}[tb]
\centering
\caption{Key Python package versions used in the implementation.}
\label{tab:packages}
\small
\begin{tabular}{ll}
\toprule
\textbf{Package} & \textbf{Version} \\
\midrule
torch        & 2.8.0 \\
torchvision  & 0.23.0 \\
transformers & 4.56.1 \\
accelerate   & 1.10.1 \\
datasets     & 4.2.0 \\
lighteval    & 0.11.0 \\
\bottomrule
\end{tabular}
\end{table}

Text-only experiments run on two NVIDIA RTX PRO 4000 GPUs with at least 24~GB of VRAM each. 
If we disable the teacher model, text-only training fits on a single GPU. Multimodal training is more resource-demanding:
the unified multimodal pipeline requires at least 48~GB of VRAM, and joint training of all parameters requires roughly 60~GB.
The multimodal experiments therefore run on two NVIDIA Tesla L40 GPUs or two NVIDIA A100 80~GB GPUs.
Throughput measurements use \texttt{llama.cpp} build \texttt{4df6e859e} (7349)~\citep{llama_cpp}.

\subsection{Runtime Configurations}
\label{sec:runtime_config}

Table~\ref{tab:gemma3_270m_configs} lists the active component dimensions for each head configuration \(k \in \{1,2,3,4\}\).

\begin{table}[tb]
\centering
\small
\setlength{\tabcolsep}{4pt}
\begin{tabular}{lcccc}
\toprule
 & H1 & H2 & H3 & H4 \\
\midrule
\(H_{\mathrm{kv}}\) & 1 & 1 & 1 & 1 \\
query heads \(k\) & 1 & 2 & 3 & 4 \\
\(d_{E,\mathrm{active}}\) & 160 & 320 & 480 & 640 \\
\(d_{\mathrm{ff},\mathrm{active}}\) & 512 & 1024 & 1536 & 2048 \\
\(d_{Q,\mathrm{active}}\) & 256 & 512 & 768 & 1024 \\
\bottomrule
\end{tabular}
\caption{HydraLLM's Gemma 3 270M runtime configurations and active component dimensions. \(d_{E,\mathrm{active}}\) is the active embedding width, \(d_{\mathrm{ff},\mathrm{active}}\) the active FFN neuron count, and \(d_{Q,\mathrm{active}}\) the total query projection dimension. MatFormer's W0.25-W1.0 have the same \(d_{\mathrm{ff},\mathrm{active}}\) as H1-H4, but otherwise keep the dimensions of H4.}
\label{tab:gemma3_270m_configs}
\end{table}

\subsection{Multimodal Dataset Composition}
\label{sec:dataset-compositions}

Tables~\ref{tab:mm_dataset_main} and~\ref{tab:mm_dataset_scaled} report the dataset mixtures for the main 2{,}500-step multimodal runs and the scaled 16{,}000-step runs, respectively. 
The scaled setting preserves the same broad mixture of captioning, question answering, code generation, and text-only data, but increases the number of samples.
The mixtures comprise PixMo-Cap~\citep{allenai_pixmocap}, our recaptioned version of COCO~\citep{coco_lin2015}, LLaVA Pretrain~\citep{liuhaotian_llava_pretrain}, PixMo Ask Model Anything~\citep{allenai_pixmo_ask_model_anything}, PixMo Cap-QA~\citep{allenai_pixmocapqa}, the VQA and code subsets of CoSyn-400K~\citep{allenai_cosyn400k}, and text-only samples from Wikipedia~\citep{wikimedia_wikipedia} and FineWeb-Edu~\citep{lozhkov2024fineweb-edu}.

\begin{table}[tb]
\centering
\caption{Composition of the multimodal dataset of the main
multimodal experiments.}
\label{tab:mm_dataset_main}
\small
\begin{tabular}{lr}
\toprule
\textbf{Dataset} & \textbf{Num Samples} \\
\midrule
\multicolumn{2}{l}{\emph{Captioning \& Pretraining}} \\
PixMo-Cap         & 50{,}000 \\
COCO Recaptioned  & 12{,}000 \\
LLaVA Pretrain    & 50{,}000 \\
\multicolumn{2}{l}{\emph{Question Answering}} \\
PixMo Ask Model Anything & 30{,}000 \\
PixMo Cap-QA             & 30{,}000 \\
CoSyn 400K VQA           & 100{,}000 \\
\multicolumn{2}{l}{\emph{Code Generation}} \\
CoSyn 400K Code          & 3{,}200 \\
\multicolumn{2}{l}{\emph{Text Only}} \\
Wikipedia EN             & 14{,}800 \\
FineWeb-Edu 10BT         & 30{,}000 \\
\midrule
Total                    & 320{,}000 \\
\bottomrule
\end{tabular}
\end{table}

\begin{table}[tb]
\centering
\caption{Composition of the multimodal dataset of the scaled
16{,}000-step multimodal training runs.}
\label{tab:mm_dataset_scaled}
\small
\begin{tabular}{lr}
\toprule
\textbf{Dataset} & \textbf{Num Samples} \\
\midrule
\multicolumn{2}{l}{\emph{Captioning \& Pretraining}} \\
PixMo-Cap         & 500{,}000 \\
COCO Recaptioned  & 100{,}000 \\
LLaVA Pretrain    & 500{,}000 \\
\multicolumn{2}{l}{\emph{Question Answering}} \\
PixMo Ask Model Anything & 125{,}000 \\
PixMo Cap-QA             & 125{,}000 \\
CoSyn 400K VQA           & 1{,}000{,}000 \\
\multicolumn{2}{l}{\emph{Code Generation}} \\
CoSyn 400K Code          & 50{,}000 \\
\multicolumn{2}{l}{\emph{Text Only}} \\
Wikipedia EN             & 200{,}000 \\
FineWeb-Edu 10BT         & 200{,}000 \\
\midrule
Total                    & 2{,}800{,}000 \\
\bottomrule
\end{tabular}
\end{table}

Projection alignment trains on a separate mixture of 64{,}000 samples from COCO Recaptioned and 64{,}000 samples from PixMo-Cap, for a total of 128{,}000 samples.
We generate COCO Recaptioned locally by recaptioning COCO training images with Qwen~3~VL~4B~Instruct \citep{qwen3vltechnicalreport} utilizing the fixed descriptive instruction ``Describe the image in 3-5 sentences!''.
The resulting captions are denser and more uniform than the original short COCO annotations.

\subsection{CLIPScore Evaluation}

The CLIPScore describes how to measure the caption quality of a model, but it does not define the underlying dataset or the CLIP-based~\cite{clip_radford2021} model~\cite{CLIPScore}. 
We use the entire COCO validation set to calculate the similarity score with clip-vit-large-patch14~\cite{clip_radford2021}.

\subsection{Training Hyperparameters}
\label{sec:appendix-hyperparameters}

Table~\ref{tab:hyperparameters} summarizes the hyperparameters for the text-only and default multimodal training runs.
The scaled multimodal runs retain the settings of the corresponding 2{,}500-step runs, but train for 16{,}000 steps on the larger dataset mixture in Table~\ref{tab:mm_dataset_scaled}, optionally train the vision encoder at a learning rate of $6\times10^{-6}$ and reduce the text model warmup from 125 to 75 steps, while, if enabled, adding 75 warmup steps for the vision encoder.
Compared with text-only training, we employ a lower text-backbone learning rate, a separate projection learning rate, a larger effective batch size, a larger knowledge-distillation weight when distilling from Gemma~3~4B, and uniform head sampling instead of the non-uniform text-only sampling weights for both multimodal runs.
The sampling weight array assigns a weight to each head for random sampling.

\begin{table*}[!ht]
\centering
\caption{Training hyperparameters for the text-only and default multimodal runs.
The full multimodal run where all components are trained differs only in vision encoder learning rate ($6\times10^{-6}$) and warmup steps (text: 75, vision: 75).
Where not listed, the multimodal configuration inherits the text default.}
\label{tab:hyperparameters}
\small
\begin{tabular}{lcc}
\toprule
\textbf{Hyperparameter} & \textbf{Text} & \textbf{Multimodal} \\
\midrule
Sequence length              & 3072        & 3072          \\
Learning rate (text backbone) & $3\times10^{-4}$ & $2\times10^{-5}$ \\
Learning rate (projection)    & --                & $2\times10^{-4}$ \\
LR schedule                   & cosine       & cosine        \\
Weight decay                  & 0.1          & 0.1           \\
Optimizer                     & AdamW        & AdamW         \\
Adam betas                    & [0.9, 0.999] & [0.9, 0.999]  \\
Adam epsilon                  & $1\times10^{-8}$ & $1\times10^{-8}$ \\
Warmup ratio (text)           & 0.01         & --            \\
Warmup steps (text model)     & --           & 125           \\
Warmup steps (projection)     & --           & 25            \\
Training steps                & 20{,}000    & 2{,}500       \\
Effective batch size          & 16           & 128           \\
Gradient clipping (norm)      & 1.0          & 1.0           \\
Head selection                & [1, 2, 3, 4] & [1, 2, 3, 4]  \\
Sampling weights              & [1, 1, 1, 3] (non-uniform) & [1,1,1,1] (uniform)       \\
Loss                          & CE + $z$-loss + KD & CE + $z$-loss + KD \\
$z$-loss alpha                & $1\times10^{-5}$ & $1\times10^{-5}$ \\
KD weight                     & 0.1          & 0.2           \\
KD temperature                & 2            & 2 \\
Teacher model                 & Gemma~3~270M & Gemma~3~4B    \\
Precision                     & bf16         & bf16 \\
Random seed                   & 42           & 42            \\
\bottomrule
\end{tabular}
\end{table*}

\section{Additional Text Experiments}
\label{sec:appendix-text}

This section presents the text training ablations, followed by the importance metrics and additional reordering results.

\subsection{Training Recipe Ablation}
\label{sec:ablations}

Table~\ref{tab:text_ablation_full} reports the full text-training ablation across ARC-Easy, HellaSwag, and Trivia\-QA.
We evaluate ARC-Easy and HellaSwag with both raw and answer-length-normalized accuracy with zero- and ten-shot prompting respectively, and Trivia\-QA with five-shot exact-match accuracy.
Relative to the default recipe, the ablations disable the teacher, use uniform head sampling, restrict training to head configurations \(\{2,3,4\}\), increase the knowledge-distillation weight to 0.3, use the FineWeb-Edu 10BT subset, apply an EMA decay of 0.999, increase the effective batch size to 128, or reduce the learning rate to \(2\times10^{-5}\).
The long-run variant instead extends training to one million steps and uses a larger FineWeb-Edu 350BT subset.

\begin{table*}[!ht]
\centering
\caption{Training recipe ablation across text benchmarks after conversion of the static Gemma 3 270M model, with MatFormer as a width-pruning comparison.
Values are accuracies in percent, with standard error, for the four runtime head configurations H1--H4 and MatFormer widths W0.25--W1.0.
ARC-Easy Norm is answer-length-normalized accuracy. Trivia\-QA is five-shot exact-match accuracy.}
\label{tab:text_ablation_full}
\scriptsize
\setlength{\tabcolsep}{3pt}
\begin{tabular}{l cccc}
\toprule
\textbf{Variant} & \textbf{H1/W0.25} & \textbf{H2/W0.5} & \textbf{H3/W0.75} & \textbf{H4/W1.0} \\
\midrule
\multicolumn{5}{l}{\emph{ARC-Easy (Raw)}} \\
MatFormer        & $53.9 \pm 1.02$ & $55.6 \pm 1.02$ & $54.6 \pm 1.02$ & $56.0 \pm 1.02$ \\
default          & $39.0 \pm 1.00$ & $46.3 \pm 1.02$ & $51.9 \pm 1.03$ & $56.4 \pm 1.02$ \\
1M steps         & $43.6 \pm 1.02$ & $51.1 \pm 1.03$ & $56.9 \pm 1.02$ & $60.3 \pm 1.00$ \\
no teacher       & $37.2 \pm 1.00$ & $46.2 \pm 1.03$ & $51.9 \pm 1.02$ & $56.1 \pm 1.02$ \\
uniform sampling & $38.8 \pm 0.99$ & $48.2 \pm 1.02$ & $52.2 \pm 1.02$ & $55.4 \pm 1.02$ \\
heads 2, 3, 4    & $26.8 \pm 0.91$ & $46.2 \pm 1.02$ & $52.3 \pm 1.03$ & $57.8 \pm 1.01$ \\
KD=0.3           & $37.9 \pm 1.00$ & $46.3 \pm 1.02$ & $52.9 \pm 1.02$ & $58.8 \pm 1.01$ \\
FW-Edu 10BT      & $38.3 \pm 1.00$ & $45.7 \pm 1.02$ & $52.6 \pm 1.02$ & $56.7 \pm 1.02$ \\
EMA=0.999        & $26.8 \pm 0.91$ & $28.3 \pm 0.92$ & $29.3 \pm 0.93$ & $54.2 \pm 1.02$ \\
batch size=128   & $31.5 \pm 0.95$ & $39.2 \pm 1.00$ & $47.3 \pm 1.02$ & $57.3 \pm 1.02$ \\
lower LR=2e-5    & $30.4 \pm 0.94$ & $34.4 \pm 0.97$ & $42.1 \pm 1.01$ & $57.4 \pm 1.01$ \\
\midrule
\multicolumn{5}{l}{\emph{ARC-Easy (Norm)}} \\
MatFormer        & $46.7 \pm 1.02$ & $49.8 \pm 1.03$ & $49.3 \pm 1.03$ & $50.7 \pm 1.03$ \\
default          & $35.8 \pm 0.98$ & $40.0 \pm 1.01$ & $46.5 \pm 1.02$ & $50.8 \pm 1.03$ \\
1M steps         & $38.6 \pm 1.00$ & $44.5 \pm 1.02$ & $49.5 \pm 1.03$ & $52.7 \pm 1.02$ \\
no teacher       & $34.4 \pm 0.99$ & $40.5 \pm 1.01$ & $46.5 \pm 1.02$ & $49.7 \pm 1.03$ \\
uniform sampling & $36.3 \pm 0.97$ & $41.5 \pm 1.01$ & $46.8 \pm 1.02$ & $49.0 \pm 1.03$ \\
heads 2, 3, 4    & $28.5 \pm 0.92$ & $41.2 \pm 1.01$ & $47.0 \pm 1.02$ & $51.0 \pm 1.03$ \\
KD=0.3           & $34.7 \pm 0.98$ & $40.2 \pm 1.01$ & $48.4 \pm 1.03$ & $51.9 \pm 1.03$ \\
FW-Edu 10BT      & $36.1 \pm 0.99$ & $39.9 \pm 1.00$ & $48.2 \pm 1.03$ & $51.8 \pm 1.03$ \\
EMA=0.999        & $27.6 \pm 0.92$ & $27.7 \pm 0.92$ & $28.8 \pm 0.93$ & $51.4 \pm 1.03$ \\
batch size=128   & $30.9 \pm 0.95$ & $34.8 \pm 0.98$ & $42.2 \pm 1.01$ & $50.2 \pm 1.03$ \\
lower LR=2e-5    & $30.8 \pm 0.95$ & $32.2 \pm 0.96$ & $37.8 \pm 1.00$ & $52.8 \pm 1.02$ \\
\midrule
\multicolumn{5}{l}{\emph{HellaSwag (Raw)}} \\
MatFormer        & $29.2 \pm 0.45$ & $30.1 \pm 0.46$ & $30.7 \pm 0.46$ & $31.3 \pm 0.46$ \\
default          & $26.2 \pm 0.44$ & $26.9 \pm 0.44$ & $29.0 \pm 0.45$ & $31.1 \pm 0.46$ \\
1M steps         & $26.3 \pm 0.44$ & $28.2 \pm 0.45$ & $30.7 \pm 0.46$ & $32.5 \pm 0.47$ \\
no teacher       & $26.0 \pm 0.44$ & $27.1 \pm 0.44$ & $28.9 \pm 0.45$ & $31.0 \pm 0.46$ \\
uniform sampling & $26.2 \pm 0.44$ & $27.1 \pm 0.44$ & $28.9 \pm 0.45$ & $29.9 \pm 0.46$ \\
heads 2, 3, 4    & $25.7 \pm 0.44$ & $27.1 \pm 0.44$ & $29.1 \pm 0.45$ & $31.2 \pm 0.46$ \\
KD=0.3           & $25.7 \pm 0.44$ & $26.9 \pm 0.44$ & $28.8 \pm 0.45$ & $31.1 \pm 0.46$ \\
FW-Edu 10BT      & $25.9 \pm 0.44$ & $26.9 \pm 0.44$ & $28.6 \pm 0.45$ & $31.2 \pm 0.46$ \\
EMA=0.999        & $25.7 \pm 0.44$ & $25.6 \pm 0.44$ & $26.1 \pm 0.44$ & $31.6 \pm 0.46$ \\
batch size=128   & $25.7 \pm 0.44$ & $26.3 \pm 0.44$ & $27.6 \pm 0.45$ & $32.1 \pm 0.47$ \\
lower LR=2e-5    & $25.5 \pm 0.43$ & $26.3 \pm 0.44$ & $27.0 \pm 0.44$ & $33.7 \pm 0.47$ \\
\midrule
\multicolumn{5}{l}{\emph{HellaSwag (Norm)}} \\
MatFormer        & $31.6 \pm 0.46$ & $33.6 \pm 0.47$ & $34.6 \pm 0.47$ & $36.1 \pm 0.48$ \\
default          & $26.4 \pm 0.44$ & $28.3 \pm 0.45$ & $31.2 \pm 0.46$ & $35.6 \pm 0.48$ \\
1M steps         & $27.4 \pm 0.44$ & $30.3 \pm 0.46$ & $34.4 \pm 0.47$ & $38.4 \pm 0.49$ \\
no teacher       & $26.2 \pm 0.44$ & $27.9 \pm 0.45$ & $30.7 \pm 0.46$ & $35.2 \pm 0.48$ \\
uniform sampling & $26.5 \pm 0.44$ & $28.2 \pm 0.45$ & $31.0 \pm 0.46$ & $33.5 \pm 0.47$ \\
heads 2, 3, 4    & $26.0 \pm 0.44$ & $28.3 \pm 0.45$ & $32.2 \pm 0.46$ & $36.4 \pm 0.48$ \\
KD=0.3           & $26.0 \pm 0.44$ & $27.8 \pm 0.45$ & $31.3 \pm 0.46$ & $35.8 \pm 0.48$ \\
FW-Edu 10BT      & $26.3 \pm 0.44$ & $27.7 \pm 0.45$ & $31.3 \pm 0.46$ & $35.7 \pm 0.48$ \\
EMA=0.999        & $26.0 \pm 0.44$ & $26.1 \pm 0.44$ & $26.3 \pm 0.44$ & $37.2 \pm 0.48$ \\
batch size=128   & $24.8 \pm 0.43$ & $27.1 \pm 0.44$ & $29.7 \pm 0.46$ & $38.0 \pm 0.48$ \\
lower LR=2e-5    & $25.5 \pm 0.43$ & $26.3 \pm 0.44$ & $29.1 \pm 0.45$ & $40.8 \pm 0.49$ \\
\midrule
\multicolumn{5}{l}{\emph{Trivia\-QA}} \\
MatFormer        & $1.47 \pm 0.09$ & $4.70 \pm 0.16$ & $5.18 \pm 0.17$ & $6.19 \pm 0.18$ \\
default          & $0.15 \pm 0.03$ & $0.23 \pm 0.04$ & $3.69 \pm 0.14$ & $5.51 \pm 0.17$ \\
1M steps         & $0.13 \pm 0.03$ & $0.66 \pm 0.06$ & $2.35 \pm 0.11$ & $3.55 \pm 0.14$ \\
no teacher       & $0.01 \pm 0.01$ & $0.02 \pm 0.01$ & $1.87 \pm 0.10$ & $2.16 \pm 0.11$ \\
uniform sampling & $0.14 \pm 0.03$ & $0.33 \pm 0.04$ & $3.38 \pm 0.13$ & $5.48 \pm 0.17$ \\
heads 2, 3, 4    & $0.00 \pm 0.00$ & $0.25 \pm 0.04$ & $3.81 \pm 0.14$ & $6.50 \pm 0.18$ \\
KD=0.3           & $0.14 \pm 0.03$ & $0.18 \pm 0.03$ & $3.72 \pm 0.14$ & $6.88 \pm 0.19$ \\
FW-Edu 10BT      & $0.11 \pm 0.02$ & $0.22 \pm 0.04$ & $2.89 \pm 0.12$ & $6.12 \pm 0.18$ \\
EMA=0.999        & $0.00 \pm 0.00$ & $0.00 \pm 0.00$ & $0.01 \pm 0.01$ & $3.90 \pm 0.14$ \\
batch size=128   & $0.00 \pm 0.00$ & $0.17 \pm 0.03$ & $2.67 \pm 0.12$ & $7.17 \pm 0.19$ \\
lower LR=2e-5    & $0.00 \pm 0.00$ & $0.02 \pm 0.01$ & $1.33 \pm 0.09$ & $8.00 \pm 0.20$ \\
\bottomrule
\end{tabular}
\end{table*}

\subsection{Importance Reordering}
\label{sec:importance-metrics}

This section describes the three importance metrics we implement to reorder embedding channels, attention heads, and FFN neurons before stochastic subnetwork training.
We compute importance scores from activation statistics collected during a forward pass over a calibration text corpus~\citep{balbay2024calibration} using forward hooks~\citep{nemotron_elastic_taghibakhshi2025}. 
Each metric ranks units independently for each component. We then apply the resulting orders to the embedding dimension, the query-head axis, and the FFN intermediate axis, as described in Section~\ref{sec:importance-reordering}.

\subsubsection{Magnitude-Based Importance}

The simplest criterion is activation magnitude \citep{nemotron_elastic_taghibakhshi2025,compactlanguagemodelspruning_muralidharan2024}.
It assumes that units with consistently large activations contribute more strongly than units with small activations.

For FFN neurons, let \(g_{b,t,j}^{(\ell)}\) denote the \(j\)-th channel of the gate-projection output in layer \(\ell\), for batch element \(b\) and sequence position \(t\). The magnitude score is:
\begin{equation}
    s_{j,\ell}^{\mathrm{mag}} = \sum_{b,t} |g_{b,t,j}^{(\ell)}|
    \label{eq:reorder_mag_mlp}
\end{equation}
For attention heads, we score the query-side projection slice. 
If \(q_{b,t,h}^{(\ell)} \in \mathbb{R}^{d_h}\) denotes the query activation of head \(h\), then:
\begin{equation}
    s_{h,\ell}^{\mathrm{mag}} = \sum_{b,t} \|q_{b,t,h}^{(\ell)}\|_2
    \label{eq:reorder_mag_attn}
\end{equation}
For embedding dimensions, we compute magnitude from the normalized hidden states at the output of the pre-layer normalization.
Let \(x_{b,t,c}^{(\ell)}\) denote embedding channel \(c\) after normalization in layer \(\ell\).
The layer-local score is:
\begin{equation}
    s_{c,\ell}^{\mathrm{mag}} = \sum_{b,t} |x_{b,t,c}^{(\ell)}|
    \label{eq:reorder_mag_embed_layer}
\end{equation}
Because we consistently slice embedding dimensions across the entire decoder, we sum layer-local scores into a single global ranking:
\begin{equation}
    s_c^{\mathrm{mag}} = \sum_{\ell} s_{c,\ell}^{\mathrm{mag}}
    \label{eq:reorder_mag_embed_global}
\end{equation}
Magnitude scoring is simple and inexpensive, but it does not account for how strongly downstream weights consume a unit's output.

\subsubsection{Variance \(\times\) Consumer Norms}

The second metric, VxC, combines activation variance \citep{variance_weinstein2022} with downstream consumer strength.
The intuition is that a unit should rank highly if it varies across inputs and if strong outgoing weights amplify its output.
Given activations \(z_i\), the empirical variance is \(\operatorname{Var}(z_i)=\mathbb{E}[z_i^2]-\mathbb{E}[z_i]^2\), where the expectation is taken over the calibration data.

For FFN neurons, VxC multiplies the gate-projection activation variance by the squared norm of the corresponding down-projection column:
\begin{equation}
    s_{j,\ell}^{\mathrm{vxc}} =
    \operatorname{Var}_{b,t}\!\left(g_{b,t,j}^{(\ell)}\right)
    \cdot
    \|W_{\mathrm{down},\ell}[:,j]\|_2^2
    \label{eq:reorder_vxc_mlp}
\end{equation}
For attention heads, we first sum the channel-wise variance within each query head:
\begin{equation}
    v_{h,\ell} =
    \sum_{d=1}^{d_h}
    \operatorname{Var}_{b,t}\!\left(q_{b,t,h,d}^{(\ell)}\right)
    \label{eq:reorder_vxc_attn_var}
\end{equation}
Then we weight this term by the squared Frobenius norm of the
corresponding output-projection slice \(W^O_{\ell,h}\):
\begin{equation}
    s_{h,\ell}^{\mathrm{vxc}} =
    v_{h,\ell} \cdot \|W^O_{\ell,h}\|_F^2
    \label{eq:reorder_vxc_attn}
\end{equation}
Several projections inside a transformer block consume the embedding channels. 
Let \(\mathcal{C}_\ell = \{W^Q_\ell, W^K_\ell, W^V_\ell, W^{\mathrm{gate}}_\ell, W^{\mathrm{up}}_\ell\}\) denote these consumers. 
The per-layer embedding score is:
\begin{equation}
    s_{c,\ell}^{\mathrm{vxc}} =
    \operatorname{Var}_{b,t}\!\left(x_{b,t,c}^{(\ell)}\right)
    \cdot
    \sum_{W \in \mathcal{C}_\ell} \|W[:,c]\|_2^2
    \label{eq:reorder_vxc_embed_layer}
\end{equation}

To obtain one global ranking, we sum the embedding scores across layers, just like we do with magnitude scoring.
In comparison to magnitude, VxC directly captures a unit's role in the computation graph. 
Units that fluctuate significantly but have weak consumption receive lower rankings than those whose variation is amplified by large outgoing weights.

\subsubsection{CETT-Based Importance}

Finally, the metric based on CETT~\citep{cett_zhang2024} measures the extent to which tail truncation affects the component output.
We adapt this idea into per-unit scores based on the norm of each unit's direct contribution to the output.

For FFN neurons, let \(h_{b,t,j}^{(\ell)}\) denote the \(j\)-th component of the input to the down projection in layer \(\ell\), and let \(W_{\mathrm{down},\ell}[:,j]\) denote the corresponding output weight vector.
The contribution norm of neuron \(j\) is:
\begin{equation}
    \begin{aligned}
        c_{b,t,j}^{(\ell)}
        &= \|h_{b,t,j}^{(\ell)} W_{\mathrm{down},\ell}[:,j]\|_2 \\
        &= |h_{b,t,j}^{(\ell)}| \cdot \|W_{\mathrm{down},\ell}[:,j]\|_2
    \end{aligned}
    \label{eq:reorder_cett_mlp_contrib}
\end{equation}
For attention heads, let \(u_{b,t,h}^{(\ell)} \in \mathbb{R}^{d_h}\)
denote the output of head \(h\) before the output projection, and let
\(W^O_{\ell,h} \in \mathbb{R}^{d_{\mathrm{model}} \times d_h}\) be the
corresponding output-projection slice. The head contribution norm is:
\begin{equation}
    c_{b,t,h}^{(\ell)} = \|W^O_{\ell,h} u_{b,t,h}^{(\ell)}\|_2
    \label{eq:reorder_cett_attn_contrib}
\end{equation}
The unnormalized CETT score is the empirical mean contribution norm. 
The normalized variant divides the contribution norm by the norm of the full component output at the same token position.
For FFN neurons, this gives:
\begin{equation}
    s_{j,\ell}^{\mathrm{cett}} =
    \mathbb{E}_{b,t}\!\left[
        \frac{c_{b,t,j}^{(\ell)}}{\|y_{b,t}^{\mathrm{FFN},(\ell)}\|_2}
    \right]
    \label{eq:reorder_cett_norm_mlp}
\end{equation}
For attention heads, this gives:
\begin{equation}
    s_{h,\ell}^{\mathrm{cett}} =
    \mathbb{E}_{b,t}\!\left[
        \frac{c_{b,t,h}^{(\ell)}}{\|y_{b,t}^{\mathrm{Attn},(\ell)}\|_2}
    \right]
    \label{eq:reorder_cett_norm_attn}
\end{equation}
The normalized form is the most relevant variant for prefix slicing because it estimates a unit's contribution relative to the total component output.
Since adaptive slicing removes the tail of a dimension, this ranking directly targets the expected relative output error incurred when we discard low-ranked units.

\subsubsection{Effect of Importance Reordering}
\label{sec:appendix-reordering-ablation}

Table~\ref{tab:reordering_ablation} complements the perplexity comparison in Section~\ref{sec:reordering} with downstream accuracy after 20{,}000 training steps.
Each row corresponds to a different combination of importance metrics for embeddings, attention heads, and FFN units.
Mag-CETT-CETT improves Trivia\-QA exact-match accuracy across all head configurations, while ARC-Easy and HellaSwag remain close to the no-reordering baseline.

\begin{table*}[!ht]
\centering
\caption{Effect of importance reordering on the four nested runtime head configurations after 20{,}000 training steps.
Values are accuracies in percent with standard error.
ARC-Easy Norm is answer-length-normalized accuracy.
Trivia\-QA is five-shot exact-match accuracy.
The no-reordering variant is the default baseline from Table~\ref{tab:text_ablation_full}.}
\label{tab:reordering_ablation}
\scriptsize
\setlength{\tabcolsep}{3pt}
\begin{tabular}{l cccc}
\toprule
\textbf{Variant} & \textbf{H1} & \textbf{H2} & \textbf{H3} & \textbf{H4} \\
\midrule
\multicolumn{5}{l}{\emph{ARC-Easy (Raw)}} \\
no reordering   & $39.0 \pm 1.00$ & $46.3 \pm 1.02$ & $51.9 \pm 1.03$ & $56.4 \pm 1.02$ \\
Mag-CETT-CETT   & $38.4 \pm 1.00$ & $46.3 \pm 1.02$ & $52.2 \pm 1.02$ & $56.6 \pm 1.02$ \\
Mag-Mag-Mag     & $37.6 \pm 0.99$ & $46.8 \pm 1.02$ & $52.9 \pm 1.02$ & $57.1 \pm 1.02$ \\
VxC-VxC-VxC     & $36.9 \pm 0.99$ & $46.8 \pm 1.02$ & $51.9 \pm 1.03$ & $56.7 \pm 1.02$ \\
Mag-no-Mag      & $38.3 \pm 1.00$ & $46.3 \pm 1.02$ & $53.2 \pm 1.02$ & $56.5 \pm 1.02$ \\
Mag-VxC-Mag     & $37.7 \pm 0.99$ & $45.4 \pm 1.02$ & $52.1 \pm 1.02$ & $56.0 \pm 1.02$ \\
\midrule
\multicolumn{5}{l}{\emph{ARC-Easy (Norm)}} \\
no reordering   & $35.8 \pm 0.98$ & $40.0 \pm 1.01$ & $46.5 \pm 1.02$ & $50.8 \pm 1.03$ \\
Mag-CETT-CETT   & $34.7 \pm 0.98$ & $41.6 \pm 1.01$ & $46.9 \pm 1.02$ & $50.5 \pm 1.03$ \\
Mag-Mag-Mag     & $33.9 \pm 0.97$ & $42.0 \pm 1.01$ & $46.9 \pm 1.02$ & $51.1 \pm 1.03$ \\
VxC-VxC-VxC     & $34.2 \pm 0.97$ & $41.4 \pm 1.01$ & $47.3 \pm 1.02$ & $51.1 \pm 1.03$ \\
Mag-no-Mag      & $36.2 \pm 0.99$ & $41.1 \pm 1.01$ & $47.3 \pm 1.02$ & $49.4 \pm 1.03$ \\
Mag-VxC-Mag     & $34.9 \pm 0.98$ & $40.7 \pm 1.01$ & $46.3 \pm 1.02$ & $51.3 \pm 1.03$ \\
\midrule
\multicolumn{5}{l}{\emph{HellaSwag (Raw)}} \\
no reordering   & $26.2 \pm 0.44$ & $26.9 \pm 0.44$ & $29.0 \pm 0.45$ & $31.1 \pm 0.46$ \\
Mag-CETT-CETT   & $26.2 \pm 0.44$ & $26.8 \pm 0.44$ & $28.9 \pm 0.45$ & $30.9 \pm 0.46$ \\
Mag-Mag-Mag     & $26.1 \pm 0.44$ & $27.0 \pm 0.44$ & $28.5 \pm 0.45$ & $31.2 \pm 0.46$ \\
VxC-VxC-VxC     & $26.1 \pm 0.44$ & $26.9 \pm 0.44$ & $28.9 \pm 0.45$ & $31.1 \pm 0.46$ \\
Mag-no-Mag      & $26.0 \pm 0.44$ & $26.8 \pm 0.44$ & $28.7 \pm 0.45$ & $31.2 \pm 0.46$ \\
Mag-VxC-Mag     & $26.1 \pm 0.44$ & $26.8 \pm 0.44$ & $28.8 \pm 0.45$ & $31.1 \pm 0.46$ \\
\midrule
\multicolumn{5}{l}{\emph{HellaSwag (Norm)}} \\
no reordering   & $26.4 \pm 0.44$ & $28.3 \pm 0.45$ & $31.2 \pm 0.46$ & $35.6 \pm 0.48$ \\
Mag-CETT-CETT   & $26.3 \pm 0.44$ & $28.0 \pm 0.45$ & $31.2 \pm 0.46$ & $36.0 \pm 0.48$ \\
Mag-Mag-Mag     & $26.3 \pm 0.44$ & $28.7 \pm 0.45$ & $31.4 \pm 0.46$ & $35.9 \pm 0.48$ \\
VxC-VxC-VxC     & $25.6 \pm 0.44$ & $28.1 \pm 0.45$ & $31.3 \pm 0.46$ & $35.7 \pm 0.48$ \\
Mag-no-Mag      & $26.2 \pm 0.44$ & $28.0 \pm 0.45$ & $31.8 \pm 0.46$ & $35.9 \pm 0.48$ \\
Mag-VxC-Mag     & $25.8 \pm 0.44$ & $28.4 \pm 0.45$ & $31.8 \pm 0.46$ & $35.6 \pm 0.48$ \\
\midrule
\multicolumn{5}{l}{\emph{Trivia\-QA}} \\
no reordering   & $0.15 \pm 0.03$ & $0.23 \pm 0.04$ & $3.69 \pm 0.14$ & $5.51 \pm 0.17$ \\
Mag-CETT-CETT   & $0.21 \pm 0.03$ & $0.82 \pm 0.07$ & $4.84 \pm 0.16$ & $6.63 \pm 0.19$ \\
Mag-Mag-Mag     & $0.14 \pm 0.03$ & $0.08 \pm 0.02$ & $3.08 \pm 0.13$ & $5.59 \pm 0.17$ \\
VxC-VxC-VxC     & $0.00 \pm 0.00$ & $0.77 \pm 0.07$ & $3.73 \pm 0.14$ & $5.43 \pm 0.17$ \\
Mag-no-Mag      & $0.11 \pm 0.02$ & $0.45 \pm 0.05$ & $4.26 \pm 0.15$ & $5.57 \pm 0.17$ \\
Mag-VxC-Mag     & $0.03 \pm 0.01$ & $0.79 \pm 0.07$ & $2.99 \pm 0.13$ & $4.60 \pm 0.16$ \\
\bottomrule
\end{tabular}
\end{table*}

\section{Additional Multimodal Results}

This section provides additional results on text recovery after multimodal training and the optimization behavior of the elastic projection during alignment.

\subsection{Text Recovery after Multimodal Training}
\label{sec:text-recovery-mm}

Table~\ref{tab:mm_triviaqa} lists Trivia\-QA accuracies after text-only continued pre-training and after subsequent multimodal or text-only fine-tuning stages.
The comparison illustrates how teacher choice and continued fine-tuning affect recovery of factual recall after the multimodal stage.
The text-only baseline is trained for one million steps. We compare it with the default multimodal run and with two text fine-tuning variants that use either Gemma~3~270M or Gemma~3~4B as the teacher.

\begin{table}[tb]
\centering
\caption{Trivia\-QA accuracy (exact match, \%) across head configurations for text-only continued pre-training and subsequent multimodal or text-only fine-tuning variants.}
\label{tab:mm_triviaqa}
\small
\setlength{\tabcolsep}{5pt}
\begin{tabular}{lcccc}
\toprule
\textbf{Variant} & \textbf{H1} & \textbf{H2} & \textbf{H3} & \textbf{H4} \\
\midrule
text, 1M steps             & $0.13$ & $0.66$ & $2.35$ & $3.55$ \\
multimodal, default        & $0.80$ & $2.16$ & $4.08$ & $6.48$ \\
text fine-tune, 270M teacher & $0.12$ & $1.10$ & $3.64$ & $5.07$ \\
text fine-tune, 4B teacher & $0.08$ & $2.13$ & $4.56$ & $7.32$ \\
\bottomrule
\end{tabular}
\end{table}

The Gemma 3 4B teacher consistently outperforms the 270M teacher at the four-head configuration, suggesting that the larger teacher transfers more factual knowledge to the student.
The improved accuracy relative to the long-run text baseline is likely due to the shift from packed context-window training to training on individual padded samples, together with the lower learning rate and larger batch size used in this stage.

\subsection{Projection Alignment Loss Curves}
\label{sec:appendix-align-projection-loss-curves}

\begin{figure}[ht]
    \centering
    \includegraphics[width=\linewidth]{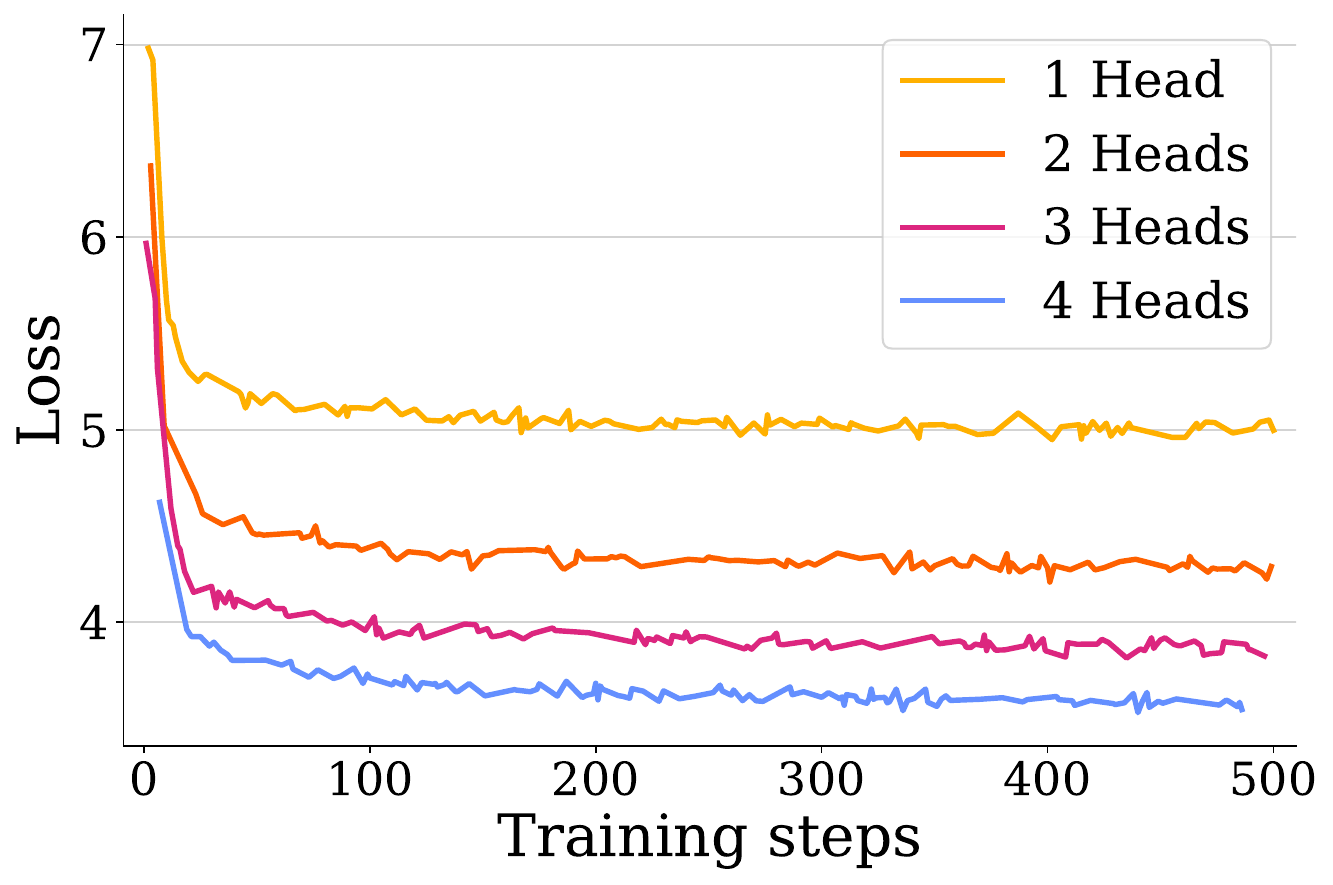}
    \caption{Training loss per-head configuration during projection alignment. The one-head model converges to a higher loss, while configurations with two or more heads reduce their loss rapidly and stabilize within the first 100 steps.}
    \label{fig:appendix-align-projection-loss}
\end{figure}

Figure~\ref{fig:appendix-align-projection-loss} shows the training loss for each head configuration during projection alignment.
The one-head configuration converges to a substantially higher loss than the wider configurations, indicating that the single-head subnetwork reaches its capacity limit earlier.
Configurations with two or more heads quickly reduce their loss and stabilize around steps 50-100, consistent with the rapid plateau observed in the main text.

\end{document}